# In situ process quality monitoring and defect detection for direct metal laser melting

Sarah Felix[2], Saikat Ray Majumder[1,*], H. Kirk Mathews[1], Michael Lexa[3], Gabriel Lipsa[4], Xiaohu Ping[1], Subhrajit Roychowdhury[1], Thomas Spears[5]

[1]*General Electric Research, Niskayuna, NY, United States,*
[2]*Rensselaer Polytechnic Institute, NY, United States,*
[3]*Systems and Technology Research, USA,*
[4]*BKW AG, Switzerland,*
[5]*Open Additive LLC, USA*

*Corresponding author. Address: GE Research, 1 Research Circle, Niskayuna, NY 12309, United States.
*Email addresses:* raymajumder@ge.com (Saikat Ray Majumder)

**Acknowledgment** This material is based upon work supported by the Air Force Research Laboratory under contract number FA8650-14-C-5702.

**Conflict of interest statement:** On behalf of all authors, the corresponding author states that there is no conflict of interest.

## Abstract

Quality control and quality assurance are challenges in Direct Metal Laser Melting (DMLM). Intermittent machine diagnostics and downstream part inspections catch problems after undue cost has been incurred processing defective parts. In this paper we demonstrate two methodologies for in-process fault detection and part quality prediction that can be readily deployed on existing commercial DMLM systems with minimal hardware modification. Novel features were derived from the time series of common photodiode sensors along with standard machine control signals. A Bayesian approach attributes measurements to one of multiple process states and a least squares regression model predicts severity of certain material defects.

# 1. Introduction

Direct metal laser melting (DMLM) is an additive manufacturing process where complex parts are built by directing a laser into a bed of powdered metal in a pattern defined by computer control. This process can exhibit significant variability from layer to layer and across the build area. The lack of sufficient quality control and assurance limits the potential of DMLM for high-performance components and high-volume production. Conventional approaches include between-build machine diagnostics and downstream part inspection, but these processes catch problems only after undue cost has been incurred processing defective parts.

The need for in-process monitoring and control has been the subject of ongoing reviews [1–4]. Recent work has explored in-process monitoring and defect detection using many different sensing and imaging modalities [5–22]. The majority of these methods employ custom instrumentation and produce high-resolution and high-speed data streams, making them difficult to deploy on commercial machines in a production setting.

This paper describes new in-process monitoring methods, developed under the guiding principle of easy deployment on original equipment manufacturers' machines with minimum modifications to available sensor packages. Two techniques are proposed, one to classify process shifts using time-dependent models of photodiode signals and another to predict severity of part defects using machine learning models.

## 2. Methods

### 2.1. Machine configuration

Figure 1 shows a representative system configuration of a commercially available DMLM machine with a standard melt pool monitoring package. An industrial PC interprets input design files to provide control to a galvanometric scanner and the laser. Additional optics direct light emitted from the melt pool to in-line sensors including photodiodes and CMOS cameras. A data acquisition system collects the sensor signals and synchronizes them with a laser trigger signal. The methods described in this paper were evaluated on Concept Laser M2 machines with QMMeltpool 3D melt pool monitoring system, which includes a down-beam photodiode and a CMOS camera, and EOS M290 machines with EOSTATE melt pool monitoring system, which includes a down-beam photodiode and an off-axis photodiode. The M2 machine used in this study comprises of a build envelope of 250 x 250 x 350 mm$^3$ (x,y,z) equipped with a laser system with power available up to 400W, a scanning speed of 7m/s and allows layer thicknesses between 20 – 80 μm. EOS M290 machine has a build envelope of 250 x 250 x 325 mm$^3$ (x,y,z), equipped with a 400 W Yb-fiber laser allowing a layer thickness of 100 μm.

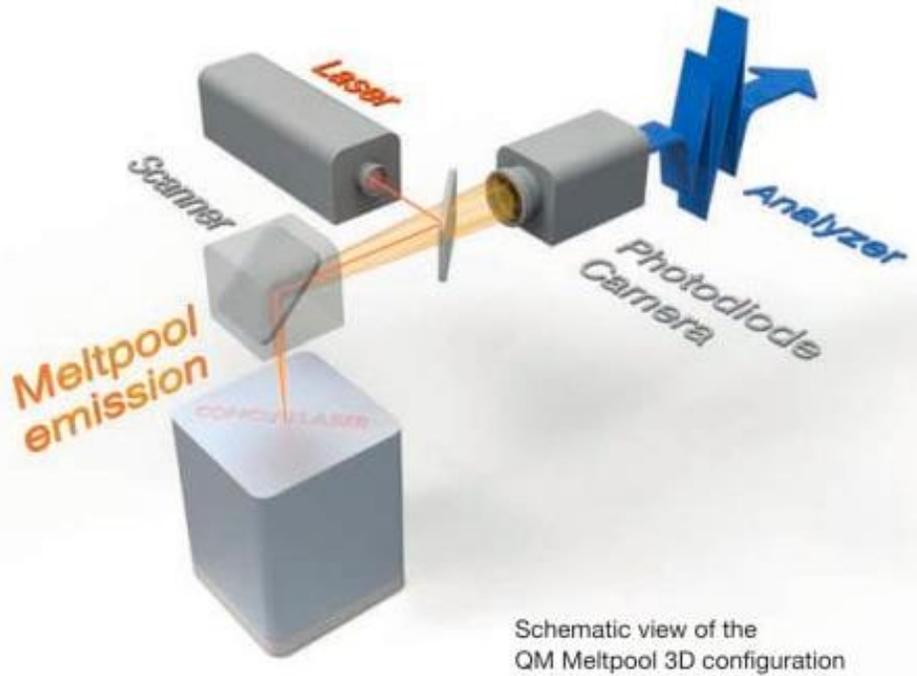

Figure 1: Schematic of a DMLM machine with a standard melt pool monitoring package. (Source: Concept Laser)

*2.2. Process fault detection*

One aim of this study was to investigate if and how the transient melt pool behavior can be used to detect off-normal process conditions. As described in Sec 2.1, the additive machines are equipped with multiple sensors (photodiodes and camera) to monitor the melt pool behavior during the build process. We consider the response of the melt pool as measured from $S$ sensors (where, S denotes the number of sensors on the machine). The sensor signals are referenced from when the laser turns on for each strike at sample $k = 1$. Samples up to $k = k_T$ are considered transient, and samples $k > k_T$ are steady state. The parameter, $k_T$, was chosen heuristically: an initial choice was based on qualitative judgement of when the process response to each laser strike settled close to a steady value for a variety of process conditions. $k_T$ was later iteratively refined based on the sensitivity of the results to this parameter, as well as computational considerations.

For process fault detection we have developed a multiple model hypothesis test (MMHT), which is a Bayesian approach that compares sensor measurements to multiple pre-estimated models to indicate a deviation from normal conditions. A model of the normal process is required, and models of one or more off-normal processes are optional. While we describe a particular form of model here, the framework is equally effective for different types of models (e.g. auto-regressive, PCA, physics-based, etc). At each sample, $k = 1, 2, ...$; and for a given process condition; the measurements from S sensors are modeled as a jointly Gaussian random vector. The set of process models (for example, processes with different shifts in laser power level), are indexed by $m = 1, 2, ..., M$. The likelihood of a set of sensor measurements $x \in \mathbb{R}^S$ is

$$p(x|m,k) = \frac{1}{\sqrt{(2\pi)^S |\Sigma_k|}} \exp\left(-\frac{1}{2}(x - \mu_k)^T \Sigma_k^{-1}(x - \mu_k)\right) \qquad 1$$

where the covariance matrix, $\Sigma_k$, and the mean vector, $\mu_k$ are constant for $k > k_T$. We further assume that measurements are uncorrelated in time which is a conservative assumption in the context of classification, and it facilitates computational flexibility.

Next, we define a kernel by a set of points aggregated in either space or time. Let $I$ be the set of N indices defining the kernel and $X_I \in \mathbb{R}^{S \times N}$ be the vector of measurements from S sensors corresponding to the indices in the kernel. Because of the independence assumption, the likelihood of $X_I$ can be computed easily for arbitrary kernels. This flexibility is important for being able to spatially aggregate measurements that may not have occurred adjacent in time. A square kernel is illustrated in the top right of Figure 2a, with indices that identify the set of measurement points within that square. We apply Bayes' rule to compute a posterior probability for each process model as

$$\Pr(m|X_I) = \frac{p(X_I|m)\Pr(m)}{\sum_{l=1}^{M+1}(p(X_I|l)\Pr(l))} \qquad 2$$

where $\Pr(m)$ is the prior probability model, $m$. The likelihood of $X_I$ is

$$p(X_I|m) = \prod_{n=1}^{N} p(x_n|m, k_n) \qquad 3$$

where $x_n$ is the $n^{th}$ column of $X_I$, and $k_n$ is the laser-on index associated with the $n^{th}$ sample in the kernel. We also introduce an "unknown" likelihood,

$$p(X_I|M+1) = p_{\text{unk}} \hspace{4cm} 4$$

where $p_{\text{unk}}$ can be tuned to prevent measurements from being attributed to an existing model if the associated likelihoods are too low. Setting the proper unknown likelihood value requires understanding the nature of the noise of the signals used to generate the empirical models for normal and fault states. Finally, we use a *maximum a posteriori* (MAP) decision rule to attribute the measurements to the model (or unknown) that has the highest posterior probability. Figure 2a illustrates the MMHT process. The left side of the figure shows an example of models for the normal process and representative faulted processes. For ease of illustration, a one-dimensional model with only on-axis photodiode data is shown, with markers and error bars representing the mean and variance, respectively, of the model for each sample time. (Note that with multiple sensors, the model would include a mean vector and covariance matrix.) The right side illustrates a square spatial kernel that aggregates points across multiple laser strikes. These points are then aggregated and analyzed as previously described.

The primary experiments were conducted on the EOS M290 system on which measurements were collected from a down- beam photodiode sensor and an off-axis photodiode at 60 kHz, for the hatch scan strategy. Raw measurements were normalized to remove fixed trends across the build plate area due to machine-specific optic artifacts. Then the measurements from both sensors were combined into a vector (i.e., $S = 2$ in Eq. 1). The mean vector and covariance matrix for the two sensors at each sample time were estimated by compiling intensity measurements from representative training builds, including a normal process and a series of processes in which laser power, scan speed, or hatch spacing were shifted in different increments. The parts in these builds included a variety of geometries such as holes, thinner walls, and sharp corners.

Finally, validation builds were run with different process parameter shifts to test the ability of the algorithm to accurately classify deviations from the normal process. In addition, builds that experienced shifted gas flow and damaged optics were considered to evaluate the detection of localized process anomalies. In the case of optic defects, the data was from a real occurrence from a production setting, so training and validation was performed on different layers of the same build.

Though the primary experiments for process fault detection analysis were performed using EOS machines, the analysis was later expanded to include Concept Laser (CL) M2 machines in a limited scope. The default scan strategy for the builds used in this analysis was to divide a part into skin and core areas. Collectively these would comprise the regions of the part that are labeled as hatch in an EOS system. Scope of our analysis has been to focus on the hatch scan strategy since the volumetric bulk of most parts are processed under those conditions. The M2 machine used in this program was of dual scan head configuration. They are situated one in front of the other with the scan head towards the back of the build plate being Head 0 (LH0) and the front scan head being Head 1 (LH1). It is important to also note that the system used here had not gone through final configuration of the sensors. Hence, in spite of holding the process constant for the entire build, there were variations in responses due to convolution of the non-configured sensors and the data being sampled near the build plate where a transient thermal environment induced by proximity to the build plate impact melt pool dynamics. This was more pronounced for the camera sensor and so for this analysis we only considered the on-axis photodiode. To classify nominal conditions for the Concept Laser parameter set, four models were built using the on-axis photodiode, one per scan head per scan parameter (skin/core): LH0 – Skin, LH0 – Core, LH1 – Skin and LH1 – Core (Figure 2b). The core parameter set uses a higher laser power than the skin parameter set and a larger spot size. The goal of this analysis was to use a part wise kernel (following the same MMHT framework) with the nominal CL models to classify parts built with the different parameter sets.

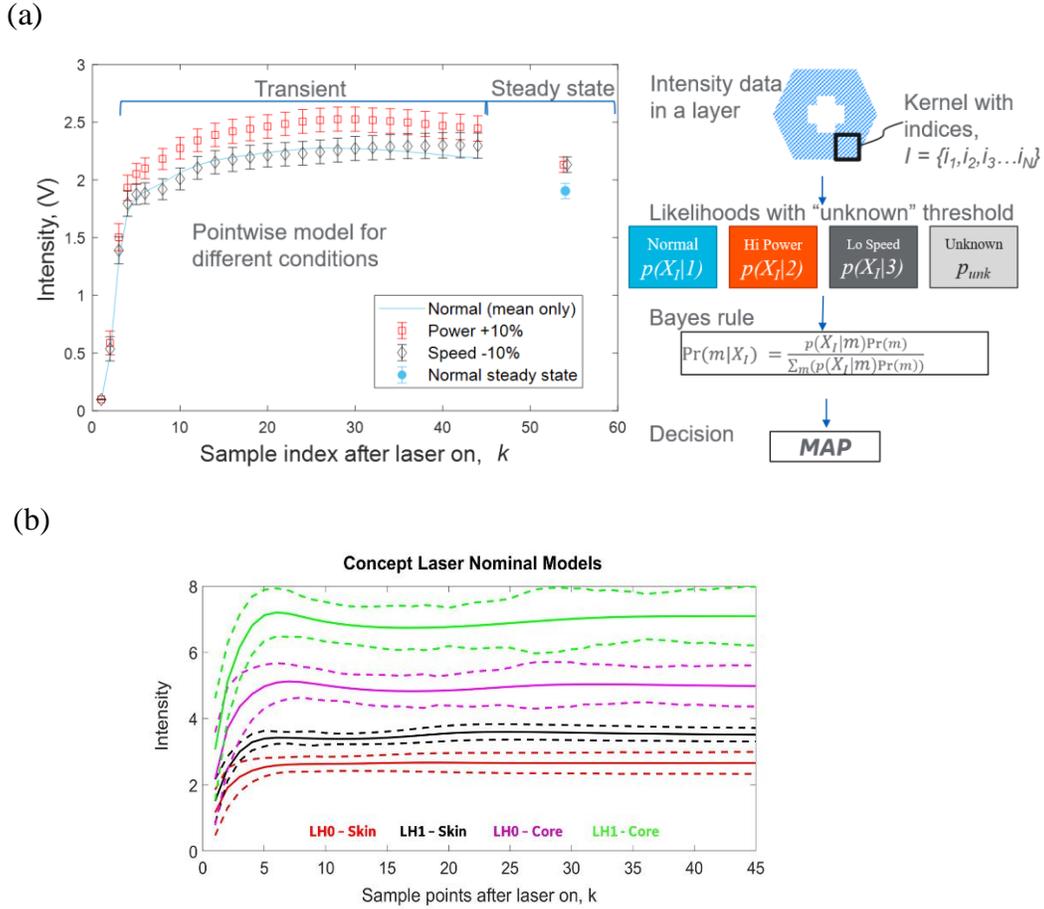

Figure 2: (a) Multiple model hypothesis test using a point-wise model of the melt pool response during laser strikes. Left side of figure shows actual models for the on-axis photodiode for the normal process and representative faulted processes. The right side illustrates a square spatial kernel that aggregates points across multiple laser strikes. These points are then analyzed as described in the text. (b) Four nominal models to analyze Concept Laser data. The solid lines represent the mean intensity models while the dashed lines represent one standard deviation bounds.

## 2.3. Part quality prediction

In addition to displaying signatures of process anomalies, melt pool behavior has a strong relationship to resultant part quality. Process deviations that cause porosity, lack of fusion, and cracks are observable in the melt pool response. Here we develop a statistical model to predict the relationship between sensor data and the severity of different material defects as measured from optical micrographs for representative pins made from CoCr. Several solid pins were built across the build plate in a Concept Laser M2 machine by varying power, speed and spot size of the processing parameters from the nominal values (details of nominal values withheld due to proprietary nature of the process parameter set) by up to 50 % to simulate defects caused by both higher and lower energy densities. A lower energy density would cause lack of fusion while a higher energy density will cause boiling porosity. Cracks are typically formed by rapid cooling of the melt pool. This expected nature of the relationship between defects and process observables is accounted for through the selection of features in the regression model, some of which include laser control parameters:

$$D_{\text{defect}} = g\left(p, s, f, \frac{p}{s}, h, \mu_{\text{part}}, \sigma_{\text{part}}, \sigma_{\mu\text{layer}}\right) \quad \quad 5$$

where $D_{\text{defect}}$ is the area fraction of a given type of defect, which can be obtained from either optical micrograph from part cut ups, CT scans of the part, or physical testing of the part. The defect score of each type is then calculated based on the ratio of the total area belonging to a certain defect class in a cross section to the total area in the said cross section. Laser parameters are given at the beginning of build process: $p$ is power, $s$ is speed, $f$ is focus, $p/s$ is linear energy density, and $h$ is hatch spacing. Sensor data are aggregated into part level variables: $\mu_{\text{part}}$ is the part-wise average intensity, $\sigma_{\text{part}}$ is the standard deviation of the part-wise intensity, and $\sigma_{\mu\text{layer}}$ is the standard deviation of layer-wise average intensity. The transfer function [Eq. 5] was obtained using stepwise ordinary least squares (OLS) optimization for each of the defects. Given the presence of collinearity, we constrained the variance inflation factor when applying stepwise OLS optimization to determine the significant model features.

### 3. Results

*3.1. Process fault detection*

We found that the transient response of the melt pool provided better discriminatory information than the steady state response. This can be seen in Figure 2a, where the steady-state melt pool intensity from a build with 10% higher power overlaps with the intensity from a build with 10% lower speed, while the transient responses to these shifts were distinct. A chosen transient length of 45 samples demonstrated compelling classification performance, while also balancing computational load.

Analysis of shifts in bulk parameters (laser power, speed, and hatch spacing) was performed using a kernel that captured all points in a layer. A square spatial kernel tuned to 0.39 mm per side was used to detect localized process anomalies such as optic damage and artifacts of low gas flow. This kernel size was tuned to balance classification accuracy and resolution.

Figure 3(a) shows the classification results for EOS M290 for bulk parameter shifts on data from validation builds that were different from the training build. There is a tradeoff of classification accuracy for an extended model set that incorporates more failure modes. In other words, with more candidate process models, there are more opportunities to confuse their output signatures. The most frequent confusion was between processes with shifted hatch spacing and a normal process, suggesting a relative insensitivity of the melt pool to this parameter shift. Another prominent confusion was between wider hatch spacing and lower power, indicating there may be a thermal effect that could be added to the model to improve classification. It might be reasonable to exclude hatch spacing in order to improve results, since this is a parameter defined by the build strategy that is unlikely to deviate during a process and it could be verified by other means upstream of the build process.

The image on the top right of Figure 3(a) shows detection of a real optic defect that occurred in a production setting. The signature for this anomalous condition was relatively dramatic, so local classification of the defect was unambiguous. The bottom right image in Figure 3(a) shows low gasflow-induced anomalies, including unknown classifications for unmodeled behavior. The model for a low gasflow defect was trained in regions of the sensor data thought to be associated with smoke occlusion of the laser beam. However, when the classification algorithm was applied, large areas of another anomalous condition were identified. This demonstrates the utility of the "unknown" model for detecting off-normal conditions, even if an explicit model of this condition does not exist.

Figure 3(b) shows results from the limited analysis that was performed on the Concept Laser M2 machine. Using the same MMHT framework, we demonstrate that models built on a combination of scan head and build parameters are highly successful in classifying parts built with different combinations of scan head and build parameters. For instance, the LH0-Skin model correctly classifies all the parts built with LH0 and Skin parameters as true positives while (correctly) identifying all the other combinations as true negatives.

The proposed fault detection methodology was optimized to complete analysis within a layer cycle time, providing an ongoing process monitoring diagnostic during a build.

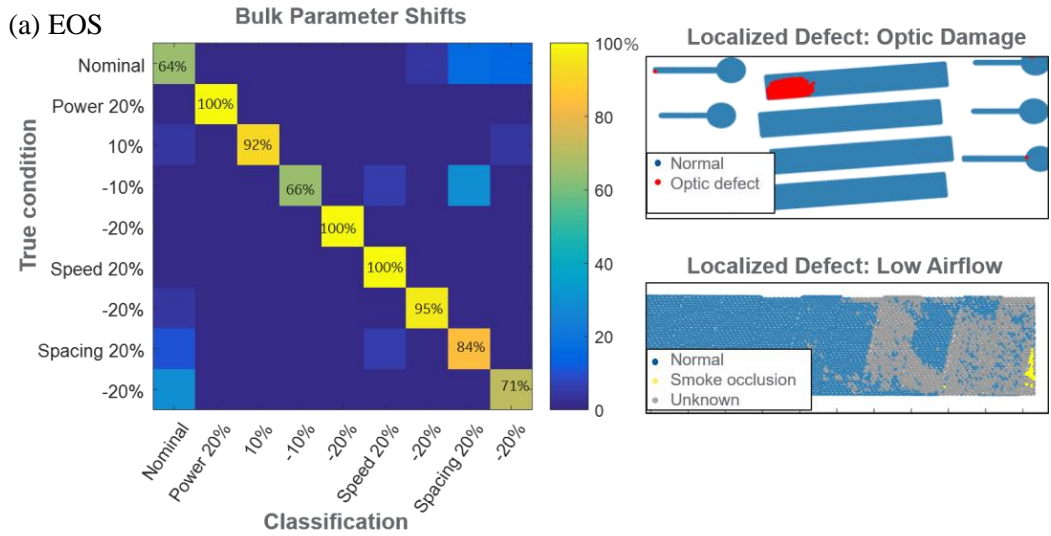

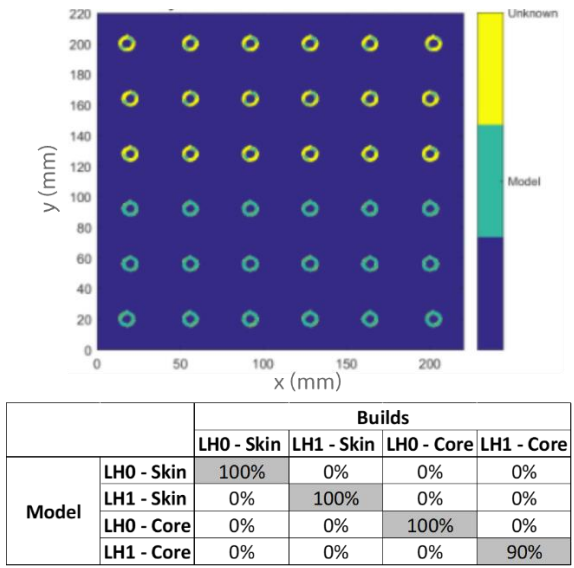

|  |  | Builds | | | |
|---|---|---|---|---|---|
|  |  | LH0 - Skin | LH1 - Skin | LH0 - Core | LH1 - Core |
| **Model** | LH0 - Skin | 100% | 0% | 0% | 0% |
|  | LH1 - Skin | 0% | 100% | 0% | 0% |
|  | LH0 - Core | 0% | 0% | 100% | 0% |
|  | LH1 - Core | 0% | 0% | 0% | 90% |

Figure 3: (a) Classification of process defects using MMHT on EOS M290. On the left is a confusion matrix for bulk parameter shifts. On the right are examples of detection of localized defects. (b) Classification of process shifts using MMHT on Concept Laser M2. The plot shows skin sections from a layer of a 36-cylinder build, where the parts built using laser head 1 were correctly classified by the LH1-skin model (green) while the parts built using laser head 0 were identified as unknown (yellow). The table shows classification results aggregated over 20 layers. Models from builds with two parameter sets (skin & core) correctly classify the parts built with corresponding parameter sets with high accuracy (>90%).

## 3.2. Part quality prediction

As previously mentioned, training builds included pins with varying parameter shifts with the intent of generating parts with correspondingly varying defect severity, defined by percent area as observed in standard optical micrographs. The micrographs of the samples are obtained from the cross-sectional view from an STL file generated by a standard micro CT module. A proprietary image processing algorithm is used on the cross-sectional views to identify pores, cracks and lack of fusion defects and get their dimensions to micron resolution. Figure 4 shows fits of the respective regression models and lists the significant factors for each one. Predictions for pores and lack of fusion had a relatively high R-Squared value, which is an indicator of goodness of fit of the model. On the other hand, cracks were harder to predict since our models were primarily trained on parts made of CoCr, a material that is typically less prone to cracking. For all defects, predictive factors were contained in both the photodiode signal features and certain laser control parameters. Part quality predictions were analyzed on a part-wise basis at the end of a build. These results demonstrate a proof of concept that for a certain degree of defects (beyond a certain threshold) caused by the variation in input parameters, part quality can be successfully predicted without deploying the time-consuming method of optical micrograph analysis.

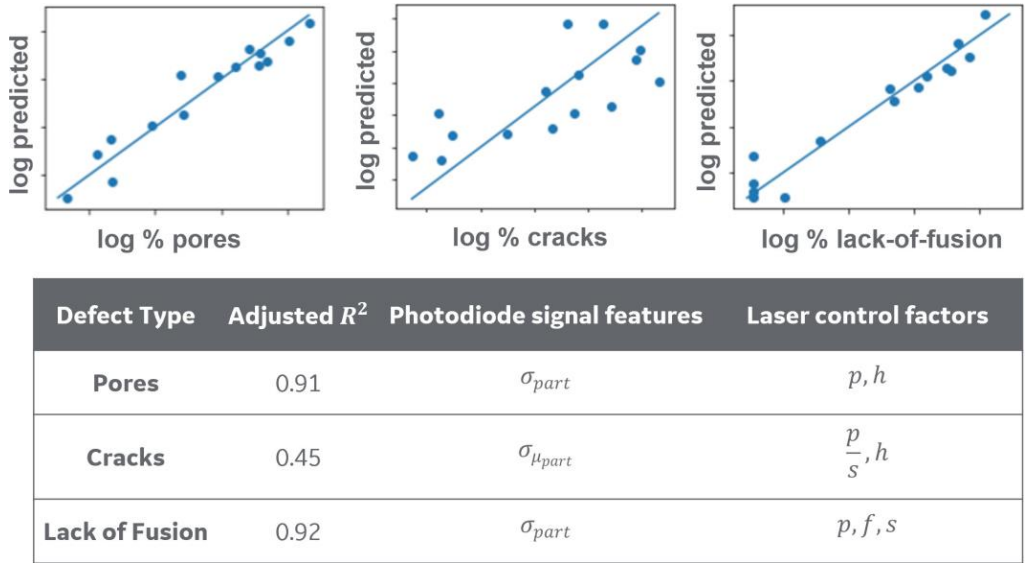

| Defect Type | Adjusted $R^2$ | Photodiode signal features | Laser control factors |
|---|---|---|---|
| Pores | 0.91 | $\sigma_{part}$ | $p, h$ |
| Cracks | 0.45 | $\sigma_{\mu_{part}}$ | $\frac{p}{s}, h$ |
| Lack of Fusion | 0.92 | $\sigma_{part}$ | $p, f, s$ |

Figure 4: Performance of defect predictor of different types of defects. A high concentration of points along the 45 degree line indicates high performance.

## 4. Conclusion
12

We successfully implemented *in situ* analytics that leverage process monitoring packages available on two commercial DMLM machines. In one experiment on an EOS system, process defect detection was able to resolve 10%-20% shifts in bulk process parameters, and detect other localized defects to sub-millimeter accuracy, all within layer cycle time. This was accomplished by extracting discriminatory information from the transient response of the melt pool in the microseconds after the laser turns on at the beginning of each strike. There are several opportunities for further refinement of this method. First, classification performance may be improved by removing the hatch spacing variation models as discussed in Section 3.1. Second, while normalization of the raw sensor measurements effectively removed gross trends due to optical artifacts, additional systematic variations were observed in the sensor measurement that limited system resolution. Ongoing work is focused on understanding these variations and creating parametric predictive models against which the sensor signals can be compared. Third, there is potential to refine performance based on a more systematic selection of the defined length of the "transient" (defined by parameter, $k_T$). Finally, incorporating other available sensor signals such as the CMOS camera in the Concept Laser system could provide process signatures with better discriminatory power.

Next steps would also address several limitations of the study described here. Detection of bulk parameter shifts was tested on builds that included a variety of part geometries *within* the build, but the training and validation builds had the same parts. Further studies are needed to characterize classification performance when the training build differs in geometry from the build being evaluated. Also, unlike the bulk parameter shift test, localized anomalies were not evaluated with all available candidate models together. A more systematic study of localized defect detection is recommended. The main challenge with local defects is that the process conditions associated with them tend to be more difficult to induce in a controlled manner in order to train reliable models.

In spite of the limitations discussed here, our results, however, show that the sensor data can be used to model nominal processes and identify in situ process shifts and also there is a high correlation between down-beam photodiode sensor data and as-built material defect severity. These early, nondestructive diagnostics enable cost savings and, with adequate validation, could potentially serve as quality assurance metrics. Note, however, that the analytics described in this letter are research tools and are not available commercially.